\documentclass{article} 
\usepackage{iclr2027_conference,times}

\usepackage{amsmath,amsfonts,bm}

\def\eqref#1{equation~\ref{#1}}

\def\1{\bm{1}}

\DeclareMathAlphabet{\mathsfit}{\encodingdefault}{\sfdefault}{m}{sl}
\SetMathAlphabet{\mathsfit}{bold}{\encodingdefault}{\sfdefault}{bx}{n}

\usepackage{hyperref}
\usepackage{url}
\usepackage{xspace}
\usepackage{graphicx}
\usepackage{makecell}
\usepackage{multicol}
\usepackage{multirow}
\usepackage{graphicx}
\usepackage{booktabs}
\usepackage{xcolor}

\def\methodName{\textcolor{black}{FeCoSplat}\xspace}
\title{\methodName: Feedback-Guided Compression for Feed-Forward 3D Gaussian Splatting}

\iclrfinalcopy

\author{
Yuxuan Li$^1$,
Yihang Chen$^{1,2}$,
Yufeng Zhang$^1$,
Jianfei Cai$^2$,
Weiyao Lin$^1$\\
$^1$Shanghai Jiao Tong University, $^2$Monash University
}

\begin{document}

\maketitle

\begin{abstract}

Feed-forward 3D Gaussian Splatting (3DGS) enables efficient novel-view synthesis from sparse multi-view images, yet its representations remain costly to store and transmit. Existing approaches compress either the input images, incurring heavy receiver-side reconstruction, or the reconstructed Gaussian primitives, which are difficult to compress due to their heterogeneous and irregular attributes. We instead compress compact intermediate features, providing a better balance between compression efficiency and receiver-side complexity.
Based on this paradigm, we propose \textbf{\methodName}, a \textbf{Fe}edback-guided \textbf{Co}mpression framework for feed-forward 3DGS. \methodName first compresses multi-view features to obtain an intermediate 3DGS, whose rendered views are used as feedback to guide a second-stage compression for further refinement. The resulting bitstreams are decoded into a compact implicit state, from which the final Gaussian primitives are reconstructed with a lightweight predictor.
Experiments demonstrate that \methodName achieves favorable rate--distortion performance, particularly at low bitrates, while requiring only 3.45M parameters for receiver-side Gaussian reconstruction. \textit{Code will be released soon}.

\end{abstract}

\section{Introduction}
\label{introduction}
3D Gaussian Splatting (3DGS)~\citep{3dgs} represents a scene with a set of anisotropic Gaussian primitives and enables efficient differentiable rendering. Conventional 3DGS obtains these primitives through time-consuming per-scene optimization. To address this limitation, feed-forward 3DGS approaches have been proposed to directly predict Gaussian representations from sparse multi-view images in a single forward pass, substantially reducing reconstruction time while generalizing across scenes~\citep{charatan2024pixelsplat,chen2024mvsplat,depthsplat,resplat,volsplat,noposplat,yonosplat,zipsplat}.

\begin{figure}[t]
    \centering
    \includegraphics[width=\linewidth]{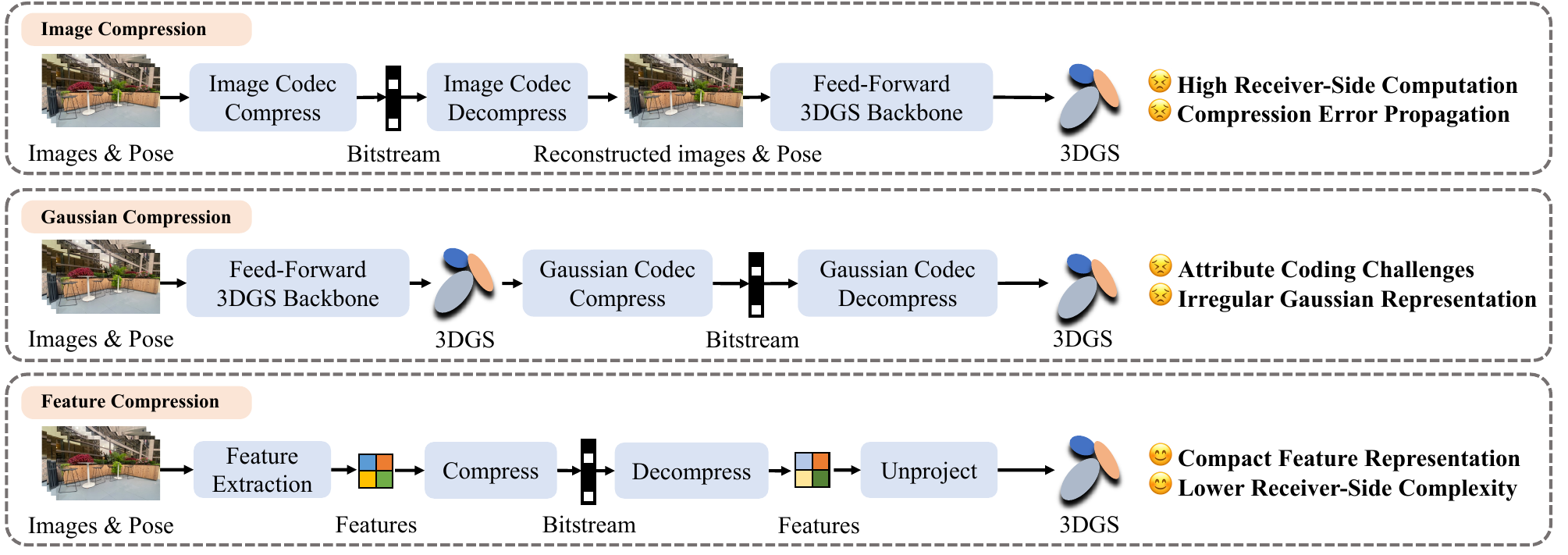}
    \caption{Different compression pipelines for feed-forward 3DGS. Image compression incurs high receiver-side computation and error propagation, while Gaussian compression is challenged by heterogeneous attributes and irregular primitives. Feature compression operates on compact, structured intermediate feature representations, balancing efficiency and receiver-side complexity.}
    \label{fig:pipeline}
\end{figure}

Despite these advances, feed-forward 3DGS approaches still produce a large number of Gaussian primitives, often requiring tens of megabytes (MB) for storage and transmission, making efficient compression essential. In a typical feed-forward 3DGS pipeline, multi-view images are first processed by a computationally intensive multi-view feature extractor to aggregate cross-view information, such as the cost-volume construction in~\citet{chen2024mvsplat}, followed by several feature transformation stages. The resulting features are then fed into lightweight Gaussian prediction heads to predict Gaussian attributes, while the camera intrinsics and extrinsics are used to determine their corresponding 3D positions. Therefore, an important preliminary question for feed-forward 3DGS compression is: \textit{what should be compressed?} 
An effective choice should achieve favorable rate--distortion (RD) performance while keeping the receiver lightweight. This is particularly important because receivers are often resource-constrained devices and may execute decoding repeatedly, whereas sender-side computation can be performed on more capable infrastructure and amortized across multiple transmissions or receivers.
Several compression paradigms can be considered for this problem, as illustrated in Figure~\ref{fig:pipeline}:

\begin{itemize}

\item \textbf{Image Compression} compresses the input multi-view images in the RGB domain using existing image codecs~\citep{elic}, and subsequently reconstructs 3DGS from the decoded images at the receiver. While this paradigm benefits from mature image compression techniques, it has two major drawbacks. First, compression distortions in the decoded images, particularly at low bitrates, can propagate through the downstream 3DGS reconstruction pipeline and degrade the final rendering quality. Second, in addition to image decoding, the receiver must execute the entire feed-forward 3DGS reconstruction pipeline, resulting in substantial receiver-side computational overhead.

\item \textbf{Gaussian Compression} takes the opposite approach by first reconstructing Gaussian primitives from the input images at the sender and then compressing their attributes for transmission. Although this avoids running the feed-forward 3DGS reconstruction pipeline at the receiver, Gaussian representations typically consist of numerous primitives with heterogeneous attributes, whose distinct distributions and dependencies make them challenging to compress effectively. This challenge becomes even more pronounced for recent feed-forward 3DGS methods~\citep{volsplat,noposplat,yonosplat,zipsplat} that produce non-pixel-aligned Gaussians with irregular spatial organization. Moreover, since these attributes directly determine the geometry and appearance of the rendered scene, compression distortions can directly degrade the final rendering quality~\citep{chen2025fcgs}.

\item \textbf{Feature Compression} provides an intermediate solution by compressing features extracted within the feed-forward 3DGS pipeline. After feature decoding, Gaussian primitives can be directly reconstructed using lightweight predictors at the receiver. Compared with explicit Gaussian representations, features are more compact, spatially structured, and more tolerant to compression distortions. Meanwhile, placing the coding interface after computationally intensive multi-view feature extraction allows the major reconstruction cost to remain at the sender, substantially reducing the computational burden at the receiver. 

\end{itemize}

We further investigate how intermediate representations can be more effectively compressed. Despite the promising potential of this paradigm, its compression efficiency remains underexplored. Specifically, existing feature compression methods such as CodecSplat~\citep{codecsplat} compress intermediate representations within a single coding stage, without explicitly considering the resulting 3D reconstruction. However, the decoded features in feed-forward 3DGS are further transformed into an explicit 3D scene representation for rendering. This reconstruction can be rendered back to the RGB observation domain, providing explicit feedback to guide subsequent coding.

To this end, we propose \methodName, a \textbf{\textit{feedback-guided compression framework for feed-forward 3DGS}}. Our key idea is to make the coding process aware of the current 3D reconstruction, rather than encoding all information in a single stage without observing the reconstruction outcome. Concretely, multi-view features are first compressed to establish an intermediate Gaussian representation, which is rendered back to the input views and compared with the original observations. The resulting reconstruction feedback reveals information that is not sufficiently captured by the first-stage representation and guides a second-stage compression for further refinement.
To support efficient information reuse across the two stages, we maintain an implicit Gaussian state in the compact feature domain. The state accumulates reconstruction-relevant information and provides decoded context for subsequent coding, allowing refinement to remain in a structured and compression-friendly representation rather than directly operating on explicit Gaussian attributes. Importantly, the intermediate Gaussian reconstruction and rendering are required only at the sender for feedback extraction. At the receiver, the transmitted bitstreams progressively update the implicit state, and only the final state is converted into explicit Gaussian primitives. This design combines reconstruction-aware compression with lightweight receiver-side reconstruction.
Our main contributions are summarized as follows:

\begin{itemize}
    \item We propose \methodName, a compression framework for feed-forward 3DGS that establishes the coding interface at compact and structured intermediate representations. This design avoids both the heavy receiver-side reconstruction required by image compression and the difficulty of efficiently compressing numerous heterogeneous Gaussian attributes.

    \item We introduce a feedback-guided two-stage coding strategy. The first stage reconstructs an 
    intermediate Gaussian representation from multi-view features, while the second stage compares its rendered results with the original observations and encodes reconstruction-conditioned feedback for refinement. A unified feature codec maintains an implicit Gaussian state represented across the two stages, which aggregates reconstruction-relevant information and serves as a compact representation for Gaussian reconstruction.

    \item Extensive experiments demonstrate that \methodName achieves improved RD performance over existing feed-forward 3DGS compression approaches while maintaining high-quality novel-view synthesis and enabling lightweight receiver-side reconstruction with only 3.45M parameters.

\end{itemize}

\section{Related Work}
\label{sec:work}

\subsection{3D Gaussian Splatting and its Feed-Forward Approaches}

3D Gaussian Splatting (3DGS)~\citep{3dgs} represents a scene as a set of anisotropic Gaussian primitives and optimizes their attributes for high-quality novel-view synthesis. 
Formally, each Gaussian is parameterized by its 3D position, opacity, scale, rotation, and spherical harmonics coefficients for view-dependent appearance.
Despite its rendering efficiency, 3DGS typically requires costly per-scene optimization. Feed-forward 3DGS addresses this limitation by directly predicting Gaussian primitives from input images, enabling generalizable novel-view synthesis. 
Splatter Image~\citep{splatter} and Flash3D~\citep{flash3d} reconstruct 3D Gaussian representations from a single image, while PixelSplat~\citep{charatan2024pixelsplat} and MVSplat~\citep{chen2024mvsplat} exploit cross-view correspondences for sparse multi-view reconstruction. DepthSplat~\citep{depthsplat} further incorporates monocular depth priors and ReSplat~\citep{resplat} iteratively refines 3D Gaussians based on rendering errors.
GS-LRM~\citep{gs-lrm} scales feed-forward Gaussian prediction with a transformer-based reconstruction model that directly predicts per-pixel Gaussians from sparse posed images.
Recent methods further improve the flexibility, quality, and efficiency of feed-forward reconstruction. NoPoSplat~\citep{noposplat}, YoNoSplat~\citep{yonosplat}, and ZipSplat~\citep{zipsplat} extend reconstruction to more flexible camera settings, while VolSplat~\citep{volsplat} and AnySplat~\citep{anysplat} improve geometric consistency through voxel-aligned prediction. Other methods explore compact representations and scalable reconstruction through efficient Gaussian allocation or latent scene representations~\citep{zpressor,sparsesplat,globalsplat,zordergs}. However, transmission-oriented compression of feed-forward 3DGS remains relatively underexplored.

\subsection{Compression of Gaussian Representations}

Existing 3DGS compression methods mainly operate directly on Gaussian primitives. 
Scene-specific approaches reduce the storage of optimized 3DGS through pruning~\citep{compact3dgs,lightgaussian,eagles,rdogaussian,ali2024trimming}, attribute quantization~\citep{compgs}, and entropy coding~\citep{chen2024hac,hac++,contextgs}.
More recent studies explore generalizable codecs that avoid per-scene codec optimization. 
FCGS~\citep{chen2025fcgs} introduces a feed-forward codec for generalizable compression of existing Gaussian representations, while D-FCGS~\citep{d-fcgs} extends feed-forward compression to dynamic Gaussian sequences by exploiting temporal correlations through motion compression and compensation. With the recent development of feed-forward 3DGS reconstruction, several methods further target the compression of feed-forward generated Gaussians. TinySplat~\citep{tinysplat} transforms Gaussian attributes into 2D representations for conventional video coding, while GenSplatCodec~\citep{gensplatcodec} introduces geometry-guided generative decoding to improve reconstruction fidelity at low bitrates. Nevertheless, these methods perform compression after explicit Gaussian reconstruction, where numerous heterogeneous attributes and irregular spatial organization make effective compression challenging.

More recently, CodecSplat~\citep{codecsplat} integrates compression into feed-forward 3DGS reconstruction by encoding intermediate features and directly reconstructing Gaussian primitives from the decoded representation. Building upon this feature compression paradigm, our method further explores reconstruction feedback as an additional cue for organizing the information derived from multi-view observations. Concretely, we decouple feature compression into two stages: the first stage of multi-view scene information for establishing an initial reconstruction and the second stage of reconstruction-conditioned feedback information for supplementing the current representation, enabling the coding process to explicitly consider the information already recovered.

\section{Method}

\subsection{Framework Overview}

\begin{figure}
    \centering
    \includegraphics[width=\linewidth]{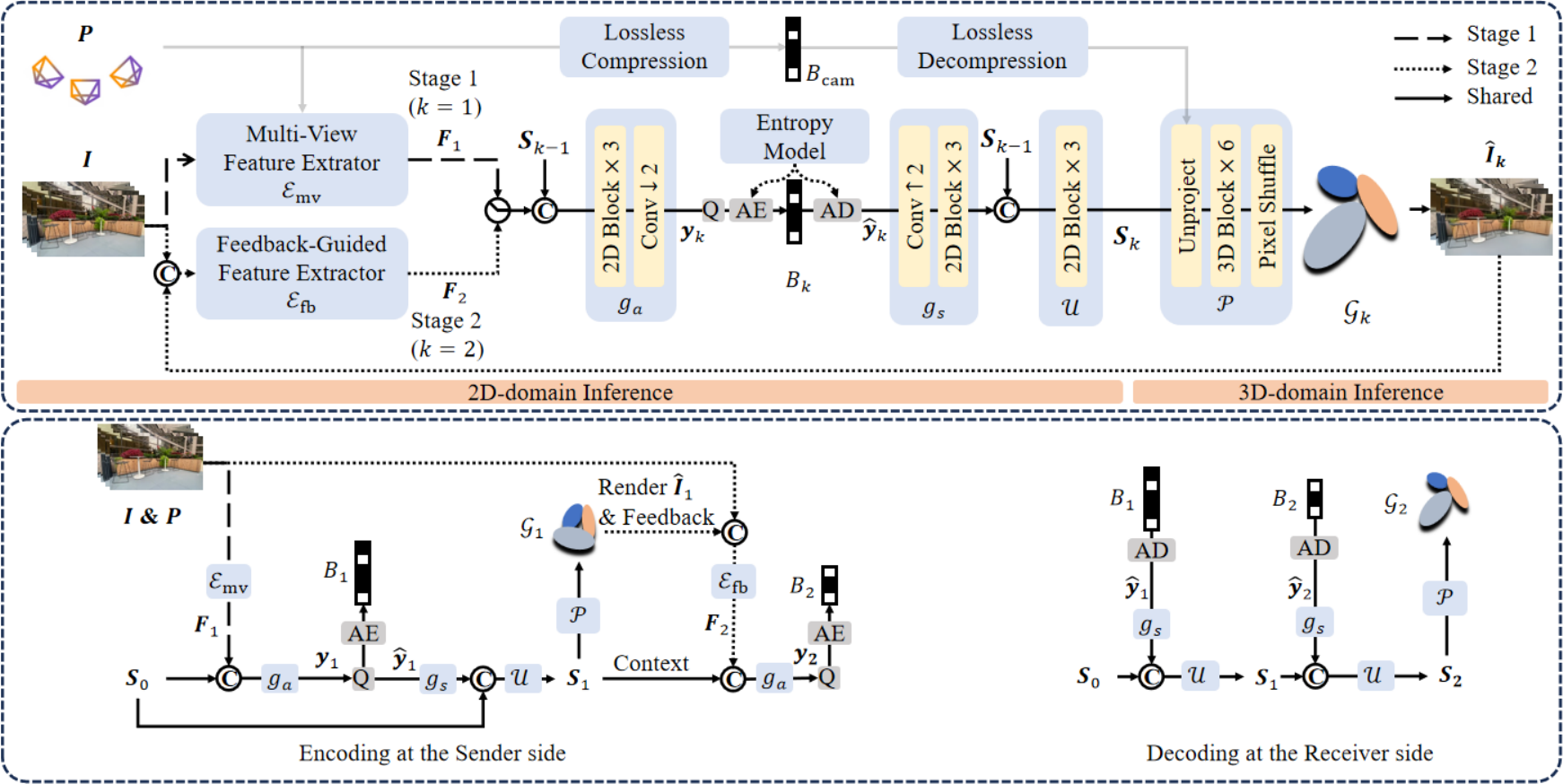}
    \caption{Overview of \methodName. \textbf{Top:} \methodName performs feature compression in two stages, encoding multi-view features in Stage 1 ($k=1$) and feedback-guided features in Stage 2 ($k=2$). A unified feature codec integrates the decoded information into an implicit Gaussian state across stages, followed by geometry-aware 3D interaction for Gaussian reconstruction. 
    \textbf{Bottom Left:} Encoding process, where reconstruction feedback is introduced to capture complementary information, and $\boldsymbol{S}_1$ serves as decoded context to reduce redundancy in the second-stage coding. \textbf{Bottom Right:} Decoding process, which requires only lightweight modules and avoids explicitly reconstructing the intermediate 3DGS. The encoding and decoding of camera parameters are omitted for clarity.}
    \label{fig:framework}
\end{figure}

As a compression-oriented framework, our approach aims at transmitting sparse multi-view observations through compact feature representations and reconstructing 3D Gaussian primitives at the receiver for novel-view rendering. 
As shown at the top of Figure~\ref{fig:framework}, given $N$ input images $\boldsymbol{I}=\{\boldsymbol{I}^i\}_{i=1}^N$ and their poses $\boldsymbol{P}=\{\boldsymbol{P}^i\}_{i=1}^N$, 
\methodName encodes them into $B=\{B_1,B_2,B_{\mathrm{cam}}\}$,
where $B_1$ and $B_2$ are feature bitstreams of two stages compressed with ANS coding~\citep{duda2009asymmetric,begaint2020compressai}
and $B_{\mathrm{cam}}$ is the camera parameters losslessly compressed using Deflate.
To generate $B_1$ and $B_2$, \methodName decouples feature compression into two stages.
The first stage extracts multi-view features $\boldsymbol{F}_1$ from the input posed images using $\mathcal{E}_{\mathrm{mv}}$ and encodes them into $B_1$, which conveys the scene information and updates the initial Gaussian state $\boldsymbol{S}_0$ to an intermediate one $\boldsymbol{S}_1$,
from which the intermediate Gaussian $\mathcal{G}_1$ is predicted.
The rendering of $\mathcal{G}_1$ is then used as reconstruction feedback to extract feedback features $\boldsymbol{F}_2$ using $\mathcal{E}_{\mathrm{fb}}$ by comparison with the input images. These features are encoded into the second-stage bitstream $B_2$ to complement the information in $\boldsymbol{S}_1$, producing the updated state $\boldsymbol{S}_2$, from which the refined Gaussian representation $\mathcal{G}_2$ is predicted. The coding processes at both the sender and receiver are shown at the bottom of Figure~\ref{fig:framework}.
The following sections detail the framework.

\subsection{Stage I: Multi-View Feature Coding}

In the first stage, geometry-aware multi-view features are extracted from the posed input images as:
\begin{equation}
\boldsymbol{F}_1
=
\mathcal{E}_{\mathrm{mv}}
\left(
\boldsymbol{I},
\boldsymbol{P}
\right),
\end{equation}
where $\mathcal{E}_{\mathrm{mv}}$ adopts the feature extraction backbone of DepthSplat~\citep{depthsplat}, consisting of a monocular transformer~\citep{depthanythingv2}, a multi-view transformer~\citep{unimatch}, and a cross-view U-Net~\citep{unet,ldm}. The monocular transformer provides strong per-view depth priors, while the multi-view transformer establishes correspondences across input views and provides matching features for cost-volume construction~\citep{chen2024mvsplat}. These features, together with the monocular geometric features and cost-volume information, are further fused by the cross-view U-Net to produce geometry-aware multi-view features $\boldsymbol{F}_1\in\mathbb{R}^{N\times C\times\frac{H}{4}\times\frac{W}{4}}$. $\boldsymbol{F}_1$ is jointly transformed with a learnable initial implicit Gaussian state $\boldsymbol{S}_0\in\mathbb{R}^{1\times C\times1\times1}$ (which is broadcast to match the spatial and view dimensions of $\boldsymbol{F}_1$) into a compact latent representation $\boldsymbol{y}_1$ using $g_a$:
$\boldsymbol{y}_1=g_a(\mathrm{concat}(\boldsymbol{F}_1, \boldsymbol{S}_0))$, where $\mathrm{concat}$ is channel-wise concat.
Specifically, to enable quantization-aware training, $\boldsymbol{y}_1$ is perturbed with additive uniform noise to approximate quantization~\citep{balle2018}, yielding $\hat{\boldsymbol{y}}_1$.
To estimate the coding rate of $\hat{\boldsymbol{y}}_1$, we employ a hyperprior~\citep{balle2018} together with a quadtree entropy model~\citep{dcvc-dc}. The hyper-latent $\hat{\boldsymbol{z}}_1$ is modeled with a factorized prior to provide side information, while the quadtree entropy model predicts the parameters of a Gaussian distribution over $\hat{\boldsymbol{y}}_1$ conditioned on $\hat{\boldsymbol{z}}_1$. The resulting likelihoods are calculated as:
\begin{equation}
\label{eq:rate}
R_1 = R(\hat{\boldsymbol{y}}_1) + R(\hat{\boldsymbol{z}}_1),
\end{equation}
which is used for end-to-end RD optimization. We refer readers to~\citet{balle2018} for further details on the entropy modeling framework.
Note that \textit{during inference}, the additive uniform noise used to approximate quantization is replaced by $\mathrm{Round}$. Then the quantized latents are entropy coded into the first-stage bitstream $B_1$. 
At the receiver, the quantized latent $\hat{\boldsymbol{y}}_1$ is directly available during training, whereas at inference time, it is recovered by entropy decoding the bitstream $B_1$. The resulting $\hat{\boldsymbol{y}}_1$  is subsequently transformed by the synthesis transform $g_s$ and incorporated into the initial state $\boldsymbol{S}_0$ through the state update module $\mathcal{U}$:
\begin{equation}
\boldsymbol{S}_1=
\mathcal{U}
\left(
\mathrm{concat}(\boldsymbol{S}_0,
g_s(\hat{\boldsymbol{y}}_1))
\right).
\end{equation}
The updated state $\boldsymbol{S}_1$ is then mapped to the intermediate 3D Gaussian representation $\mathcal{G}_1$ through the Gaussian predictor $\mathcal{P}$:
\begin{equation}
\mathcal{G}_1=\mathcal{P}\left(\boldsymbol{S}_1\right).
\end{equation}
To maintain low reconstruction complexity, $\mathcal{P}$ operates primarily on a low-resolution 2D feature grid. It first predicts per-view depth maps from $\boldsymbol{S}_1$ and back-projects the features into 3D space. Complementing the preceding 2D feature processing, KNN attention~\citep{zhao2021point} performs geometry-aware aggregation directly in 3D by collecting spatially neighboring features across views. A lightweight prediction head then estimates the Gaussian attributes, which are rearranged by PixelShuffle to produce pixel-aligned 3D Gaussians $\mathcal{G}_1$ at half the input resolution, corresponding to one Gaussian per $2\times2$ image region. This design combines efficient 2D processing with explicit 3D geometric interaction while avoiding costly high-resolution feature computation.

\subsection{Stage II: Feedback-Guided Feature Coding}

The first stage produces an intermediate Gaussian representation $\mathcal{G}_1$, which provides an initial reconstruction of the scene. 
To enable reconstruction-aware refinement, we render $\mathcal{G}_1$ back to the input viewpoints with the corresponding camera parameters $\boldsymbol{P}$:
\begin{equation}
\hat{\boldsymbol{I}}_1=
\operatorname{Render}
\left(
\mathcal{G}_1,
\boldsymbol{P}
\right).
\end{equation}
where $\hat{\boldsymbol{I}}_1$ denotes the rendered observations of $\mathcal{G}_1$ at the input viewpoints, providing explicit feedback on the current reconstruction state of the intermediate 3DGS. 
Afterwards, we jointly process each original image and its corresponding rendering using a feedback feature extractor $\mathcal{E}_{\mathrm{fb}}$:
\begin{equation}
\boldsymbol{F}_2
=
\left\{
\mathcal{E}_{\mathrm{fb}}
\left(
\operatorname{concat}
\left(\boldsymbol{I}^i,\hat{\boldsymbol{I}}_1^i\right)
\right)
\right\}_{i=1}^{N}.
\end{equation}
where $\mathcal{E}_{\text{fb}}$ applies PixelUnshuffle to the input and extracts low-resolution feedback features $\boldsymbol{F}_2\in\mathbb{R}^{N\times C\times\frac{H}{4}\times\frac{W}{4}}$.
Conditioned on the input images ${\boldsymbol{I}}$,
$\boldsymbol{F}_2$ captures information that is not sufficiently represented by $\mathcal{G}_1$.
The extracted feature representation $\boldsymbol{F}_2$ is jointly transformed with the current implicit Gaussian state $\boldsymbol{S}_1$, which serves as decoded context, into a compact latent representation:
$\boldsymbol{y}_2=g_a(\mathrm{concat}(\boldsymbol{F}_2,\boldsymbol{S}_1))$.
Following the same coding procedure as in the first stage, $\boldsymbol{y}_2$ and its corresponding hyper latent $\boldsymbol{z}_2$ follow the same quantization and entropy-modeling scheme to obtain the second-stage rate $R_2$, and are entropy coded into the second-stage bitstream $B_2$ during inference.
Then the quantized latent $\hat{\boldsymbol{y}}_2$ is decoded and integrated into the current state $\boldsymbol{S}_1$ through the shared state update module $\mathcal{U}$:
\begin{equation}
\boldsymbol{S}_2=
\mathcal{U}
\left(\mathrm{concat}(
\boldsymbol{S}_1,
g_s(\hat{\boldsymbol{y}}_2))
\right).
\end{equation}
The updated state $\boldsymbol{S}_2$ integrates the information decoded from both coding stages. Notably, the second-stage refinement is performed entirely in the compact feature-state domain rather than directly on explicit Gaussian primitives, preserving the structured and compression-friendly representation throughout the coding process. The final state is then mapped to the Gaussian representation through the shared Gaussian predictor $\mathcal{P}$:
\begin{equation}
\mathcal{G}_2=
\mathcal{P}
\left(
\boldsymbol{S}_2
\right).
\end{equation}

\textbf{Feedback-Free Decoding.} Importantly, reconstruction feedback is required only during encoding, but not during decoding, as shown at the bottom of Figure~\ref{fig:framework}. At the sender, $\boldsymbol{S}_1$ is mapped to the intermediate Gaussian representation $\mathcal{G}_1$, which is rendered to extract the feedback feature $\boldsymbol{F}_2$ for second-stage encoding. At the receiver, however, neither $\mathcal{G}_1$ nor its rendering needs to be reconstructed. Once $\boldsymbol{S}_1$ and the second-stage latent $\hat{\boldsymbol{y}}_2$ are decoded, they can be directly integrated to obtain the final state $\boldsymbol{S}_2$, from which the final Gaussian representation $\mathcal{G}_2$ is reconstructed. Therefore, the feedback loop is used only at the sender, while decoding proceeds directly in the compact feature-state domain. This makes the receiver-side decoding process causal and efficient, as each stage depends only on previously decoded states and the current bitstream, without requiring any backward feedback or intermediate rendering.

\subsection{Training Strategy}

\textbf{Loss function.} We train the two-stage framework end-to-end using a joint RD loss:
\begin{equation}
    \mathcal{L}
    =
    \sum_{k=1}^{2}
    \left(\lambda R_{k}
    +
    \mathcal{L}_{\mathrm{render},k}\right),
\end{equation}
where $R$ is the rate term, $\mathcal{L}_{\mathrm{render}}$ is the rendering loss, and $\lambda$ controls the RD trade-off.
The rendering loss $\mathcal{L}_{\mathrm{render}}$ follows the setting of~\citet{depthsplat,codecsplat}:
\begin{equation}
    \mathcal{L}_{\mathrm{render},k}
    =\sum_{i=1}^{M}(
    \mathcal{L}_{\mathrm{MSE}}(
    \boldsymbol{I}^{\mathrm{tgt}, i}
    ,
    \hat{\boldsymbol{I}}_k^{\mathrm{tgt}, i})
    +
    \beta\,\mathcal{L}_{\mathrm{LPIPS}}(
    \boldsymbol{I}^{\mathrm{tgt}, i}
    ,
    {\hat{\boldsymbol{I}}_k^{\mathrm{tgt}, i}}
    )),
\end{equation}
where $\boldsymbol{I}^{\mathrm{tgt}}$ denotes the target image, $\hat{\boldsymbol{I}}^{\mathrm{tgt}}$ is the rendering from $\mathcal{G}$, and $M$ is the number of target views. We set $\beta=0.05$.
To enable variable-rate coding, we condition each 2D block on $\lambda$ through Adaptive Layer Normalization (AdaLN)~\citep{qarv}. 
Specifically, each 2D block employs a lightweight modulation network $\phi$ to predict the modulation parameters $\boldsymbol{\gamma}$ and $\boldsymbol{\beta}$ from 
$\lambda$:
\begin{equation}
    (\boldsymbol{\gamma},\boldsymbol{\beta})
=
\phi(\log\lambda),
\end{equation}
 
Given an intermediate feature $\boldsymbol{X}$, the normalized feature is modulated as
\[
\operatorname{AdaLN}(\boldsymbol{X},\lambda)
=
\boldsymbol{\gamma}\odot\operatorname{LN}(\boldsymbol{X})
+
\boldsymbol{\beta},
\]
where $\operatorname{LN}$ denotes Layer Normalization and $\odot$ denotes element-wise multiplication. In this way, by varying $\lambda$ during training, the feature transformation is adaptively modulated according to the desired RD trade-off, enabling a single codec to operate across multiple rate points.

\section{Experiments}
\label{sec:exp}

\subsection{Implementation Details}

\textbf{Datasets.}
Following prior works on feed-forward 3D Gaussian Splatting, we conduct experiments on RealEstate10K~\citep{zhou2018re10k} and DL3DV~\citep{liang2024dl3dv}. For fair comparison, we adopt the same experimental settings as CodecSplat~\citep{codecsplat}. Specifically, we use two input views at a resolution of $256\times256$ for RealEstate10K and eight input views at $256\times448$ for DL3DV. To further assess the model's cross-dataset generalization ability, we directly evaluate our RealEstate10K-trained model on ACID~\citep{acid}, without any additional fine-tuning.

\noindent\textbf{Training Details.}
We optimize \methodName using AdamW~\citep{adamw} with an initial learning rate of $2\times10^{-4}$ and a cosine learning-rate scheduler.
The multi-view feature extractor is initialized from the pretrained DepthSplat checkpoint, while the remaining modules are trained from scratch.
To support variable-rate compression, the rate-distortion trade-off parameter $\lambda$ is sampled from a log-uniform distribution during training, with ranges of $[10^{-5}, 10^{-2}]$ and $[10^{-5}, 10^{-3}]$ for RE10K and DL3DV, respectively. All models are trained for 300K steps on four NVIDIA RTX L40S GPUs with a total batch size of 4 using BF16 mixed precision.

\begin{table}[t]
\centering
\caption{Comparison of sender/receiver parameters, coding latency, and peak GPU memory on RealEstate10K. \methodName achieves a lightweight receiver with only 3.45M parameters and 0.117s latency, while also maintaining low sender-side complexity and the lowest peak GPU memory among the compared learning-based approaches.}
\vspace{2mm}
\label{tab:params}
\begin{tabular}{lccccc}
\toprule
\multirow{2}{*}{Methods}
& \multicolumn{2}{c}{Params (M) $\downarrow$}
& \multicolumn{2}{c}{Latency (s) $\downarrow$}
& \multirow{2}{*}{\shortstack{Peak GPU Mem (GiB) $\downarrow$}} \\
\cmidrule(lr){2-3}
\cmidrule(lr){4-5}
& Sender & Receiver
& Sender & Receiver
& \\
\midrule

\textcolor{gray!90}{\textit{Image Compression}}
& & & & & \\
ELIC$\rightarrow$DepthSplat
& 26.45 & 62.40 & 0.077 & 0.181 & 1.505 \\
ELIC$\rightarrow$ReSplat
& 26.45 & 100.69 & 0.074 & 0.305 & 2.119 \\

\midrule
\textcolor{gray!90}{\textit{Gaussian Compression}}
& & & & & \\
DepthSplat$\rightarrow$TinySplat
& 38.34 & N/A\textsuperscript{*} & 1.285 & 0.375 & 1.365 \\
DepthSplat$\rightarrow$FCGS
& 40.87 & 2.05 & 2.035 & 0.821 & 4.706 \\
ReSplat$\rightarrow$TinySplat
& 76.63 & N/A\textsuperscript{*} & 1.448 & 0.497 & 1.925 \\
ReSplat$\rightarrow$FCGS
& 79.16 & 2.05 & 2.319 & 0.784 & 5.157 \\

\midrule
\textcolor{gray!90}{\textit{Feature Compression}}
& & & & & \\
CodecSplat
& 181.73 & 45.00 & 0.189 & 0.138 & 2.065 \\
\methodName
& 37.28 & 3.45 & 0.168 & 0.117 & 1.316 \\

\bottomrule
\end{tabular}

\vspace{1mm}
\begin{minipage}{\linewidth}
\footnotesize
\raggedright
\textsuperscript{*} TinySplat employs the traditional image/video codec HEVC~\citep{hevc} for encoding and decoding, and thus has no trainable codec parameters.
\end{minipage}
\end{table}

\textbf{Baselines.}
For compression performance, we compare with three paradigms introduced in section~\ref{introduction}: \textit{\textbf{Image Compression}}, \textit{\textbf{Gaussian Compression}}, and \textit{\textbf{Feature Compression}}. For Image Compression, we use ELIC~\citep{elic} to compress the input views, followed by DepthSplat~\citep{depthsplat} or ReSplat~\citep{resplat} for Gaussian reconstruction. For Gaussian Compression, Gaussian primitives reconstructed by DepthSplat or ReSplat are compressed using FCGS~\citep{chen2025fcgs} or TinySplat~\citep{tinysplat}. For Feature Compression, we compare with CodecSplat~\citep{codecsplat}, which directly compresses intermediate features for Gaussian reconstruction. For a fair comparison, we select model variants for each baseline with parameter counts as close as possible to ours, thereby maintaining comparable model capacity across different compression paradigms. 

\noindent\textbf{Evaluation Metrics.}
We evaluate reconstruction quality using LPIPS~\citep{lpips}, PSNR, and SSIM~\citep{ssim}, and measure the compression rate by the total number of bits required to represent the transmitted data. 
We further compare the computational complexity of different compression pipelines on RealEstate10K under the two-view $256\times256$ setting. We report the number of parameters, encoding and decoding latency, and peak GPU memory. All runtime measurements are conducted on the same hardware under identical settings.

\subsection{Parameters and Coding Latency}

\begin{figure}
    \centering
    \includegraphics[width=\linewidth]{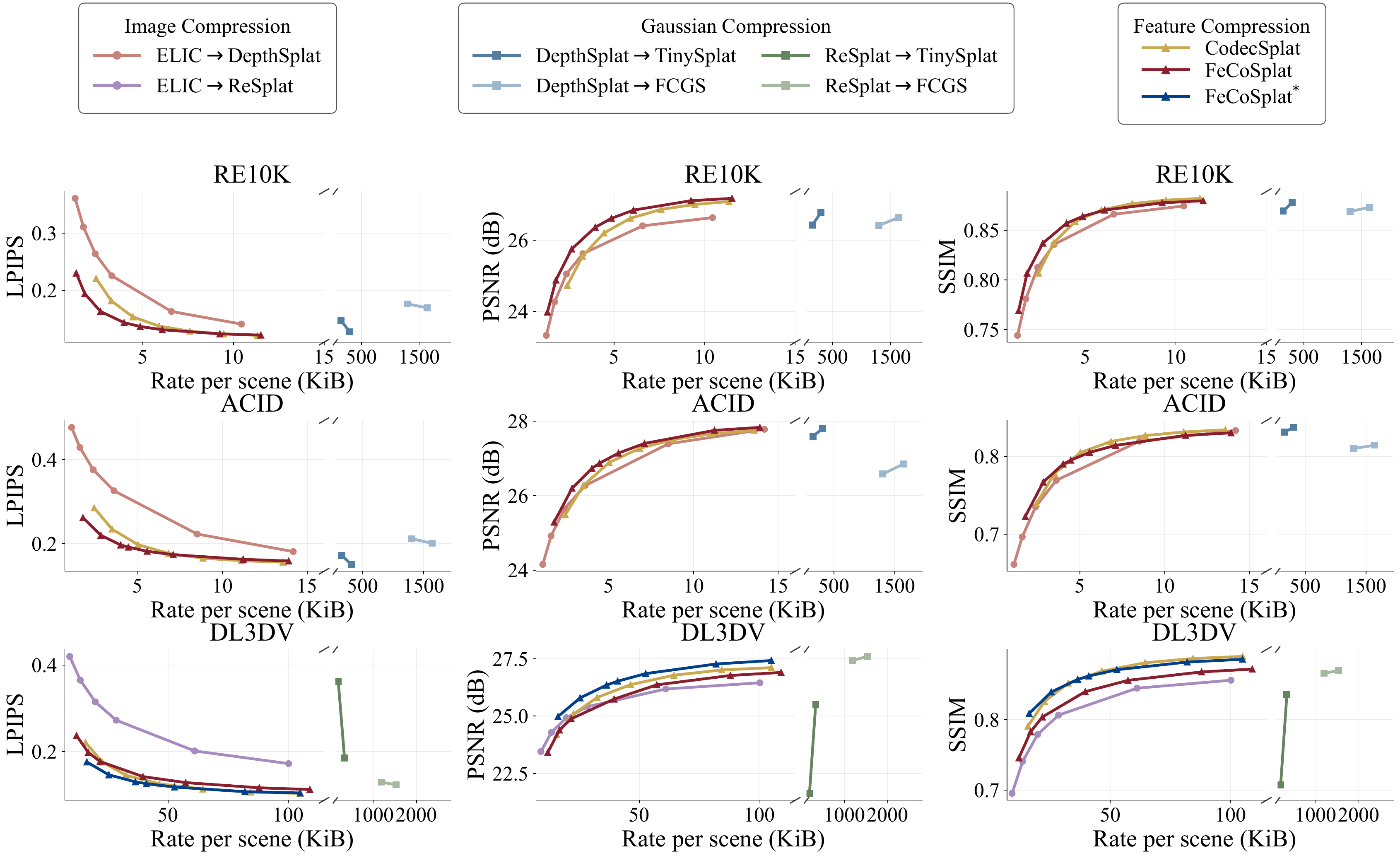}
     \caption{Performance comparison.}
    \label{fig:rd-performance}
\end{figure}

Table~\ref{tab:params}
compares the parameter counts and coding latency of different methods. 
The sender-side parameter count of \methodName is comparable to that of the Image Compression baselines~\citep{elic}, while the receiver contains only 3.45M parameters, substantially fewer than the tens to hundreds of millions of parameters required by competing approaches.
Meanwhile, \methodName achieves efficient sender- and receiver-side processing, with latencies of 0.168s and 0.117s, respectively, both lower than those of CodecSplat.
The lightweight receiver and efficient coding pipeline make \methodName particularly suitable for edge devices with limited computational resources.

\subsection{Rate-Distortion Performance}

\begin{figure}
    \centering
    \includegraphics[width=\linewidth]{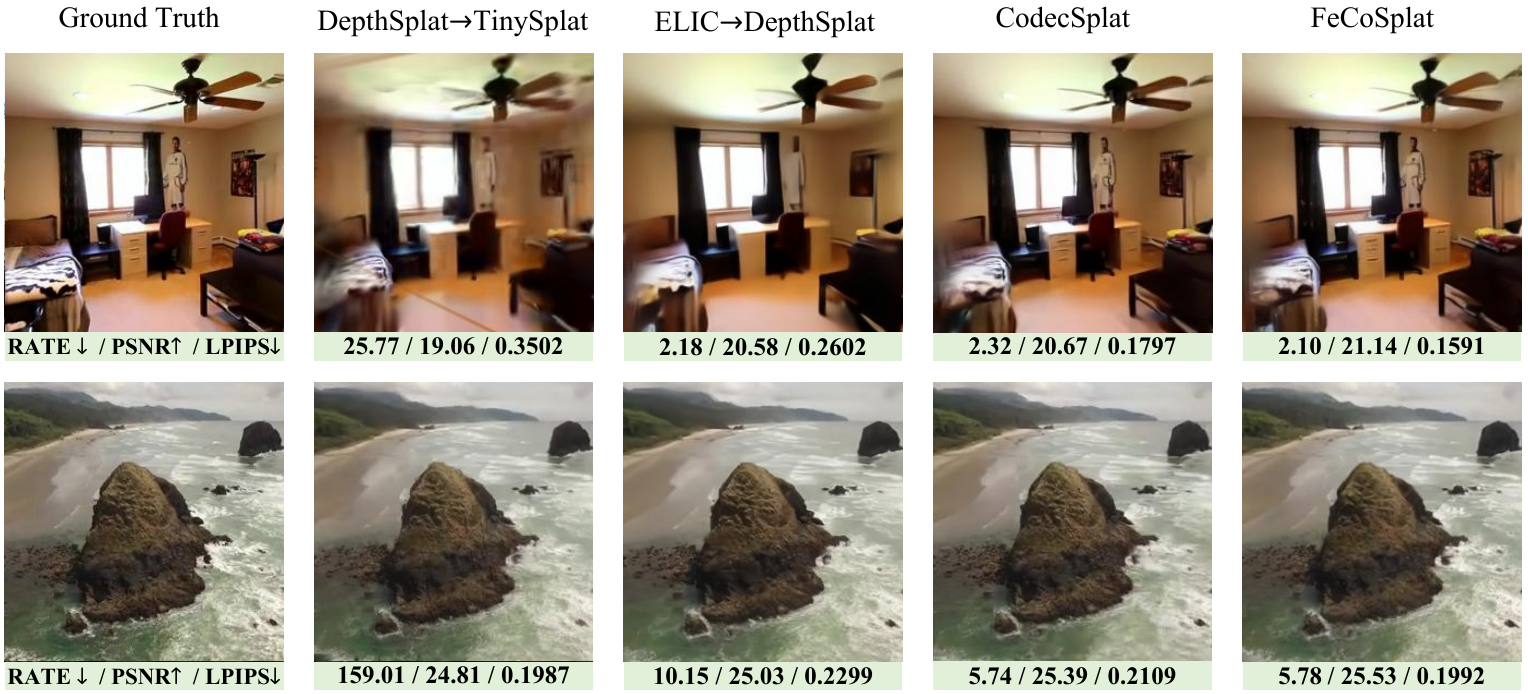}
    \caption{Qualitative comparison on RE10K and ACID. The rate per scene (KiB), PSNR (dB), and LPIPS are reported for each result. \methodName achieves favorable reconstruction quality even at extremely low bitrates compared with competing methods. More results are in Appendix~\ref{sec:visualization}.}
    \label{fig:visualization}
\end{figure}

Figure~\ref{fig:rd-performance}
shows the RD performance of \methodName against competing approaches.
On RE10K and ACID, \methodName consistently achieves the state-of-the-art RD performance, with particularly pronounced gains in LPIPS, highlighting its advantage in preserving perceptual quality under constrained bit budgets. Image-compression-based approaches degrade rapidly at low bitrates, as the limited coding budget is primarily used to preserve RGB appearance rather than the geometric information required for accurate 3D reconstruction. Gaussian-compression-based approaches exhibit considerably higher compression size, as the numerous heterogeneous Gaussian attributes and their irregular spatial organization are inherently more difficult to compress effectively.
Importantly, \methodName uses only about 20\% of CodecSplat's sender-side parameters (37M vs. 181M). 
To evaluate the scalability of our framework, we additionally construct a variant \methodName\textsuperscript{*} that adopts a larger multi-view feature extractor from CodecSplat~\citep{codecsplat}, which increases the sender-side parameters to 126.11M, while still keeping the receiver unchanged, as shown in the DL3DV results. This variant surpasses CodecSplat in RD performance, showing that \methodName can benefit from stronger feature extractors without increasing receiver-side complexity. We also compare our standard lightweight model against larger variants of image-compression-based methods. Additional results are in Appendix~\ref{appendix:exp}.

\subsection{Ablation Study}

\begin{figure}
    \centering
    \includegraphics[width=\linewidth]{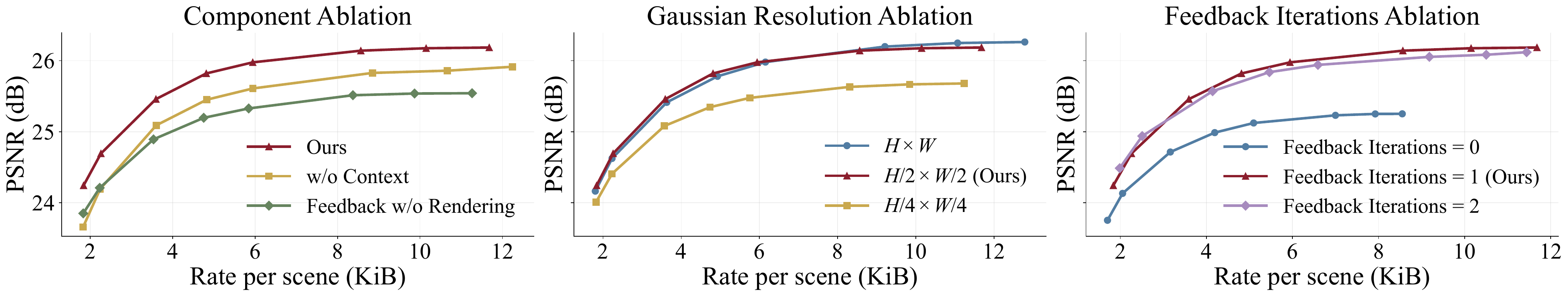}
    \caption{Ablation study on component design, Gaussian resolution and feedback iterations.}
    \label{fig:ablation}
\end{figure}

We perform ablation studies on the proposed framework. The results are shown in Figure~\ref{fig:ablation}. All variants are trained on RE10K for 100K steps under the same training settings. 
\textbf{Left:} Without the decoded context $\boldsymbol{S}_1$, the model requires a substantially larger rate budget to achieve comparable reconstruction quality, 
demonstrating the importance of exploiting decoded context across coding stages.
Removing the rendered-image feedback primarily lowers the achievable reconstruction fidelity, with the degradation becoming more remarkable at high bitrates.
\textbf{Mid:} Using an overly low Gaussian resolution significantly limits the reconstruction fidelity ceiling. Full-resolution Gaussians provide only slight quality gains at the highest bitrates, while incurring substantially higher computational cost. This indicates that a moderate Gaussian resolution provides a better trade-off.
\textbf{Right:} Without iterative refinement, the model lacks reconstruction feedback and therefore reaches a lower fidelity ceiling. Increasing the number of iterations slightly improves reconstruction quality, but the gains quickly diminish, while each additional iteration requires transmitting a new bitstream for the newly extracted information. This suggests that a small number of feedback-guided refinement stages provides a better trade-off between reconstruction fidelity and coding overhead. More discussions and visualizations regarding iteration steps are in 
Appendix~\ref{appendix:refinement}
and 
Appendix~\ref{sec:visualization}, respectively.

\section{Conclusion}

In this work, we present \methodName, a two-stage feedback-guided feature compression framework for feed-forward 3D Gaussian Splatting. \methodName progressively compresses geometry-aware multi-view features and reconstruction-conditioned feedback features, where intermediate Gaussian renderings guide the extraction of complementary information in the second stage. A shared variable-rate feature codec and an implicit Gaussian state enable efficient information reuse and progressive reconstruction without explicitly generating intermediate Gaussians at the decoder. Experimental results demonstrate favorable RD performance, particularly at low bitrates, while maintaining low decoding complexity and high-quality novel-view synthesis.

\subsection*{AI use statement}

In this work, we used generative AI tools for language editing, including grammar correction, stylistic refinement, and improving the clarity and readability of the manuscript. We did not rely on generative AI tools to generate experimental results, fabricate citations, or make scientific claims without verification. All AI-assisted text was carefully reviewed and revised by the authors to ensure technical accuracy, consistency with the actual methodology and experiments, and appropriate attribution of prior work. We take responsibility for the final content of this work, including text, claims, or artifacts produced with the aid of generative AI.

\bibliography{iclr2027_conference}
\bibliographystyle{iclr2027_conference}

\newpage

\appendix
\setcounter{figure}{0}
\setcounter{table}{0}

\renewcommand{\thefigure}{A\arabic{figure}}
\renewcommand{\thetable}{A\arabic{table}}

\begin{center}
    {\Large\bfseries Appendix}
\end{center}
\vspace{1em}

\section{More Implementation Details}

\subsection{Architectural Details of Blocks and Feature Extractors}

\begin{figure}[h]
    \centering
    \includegraphics[width=\linewidth]{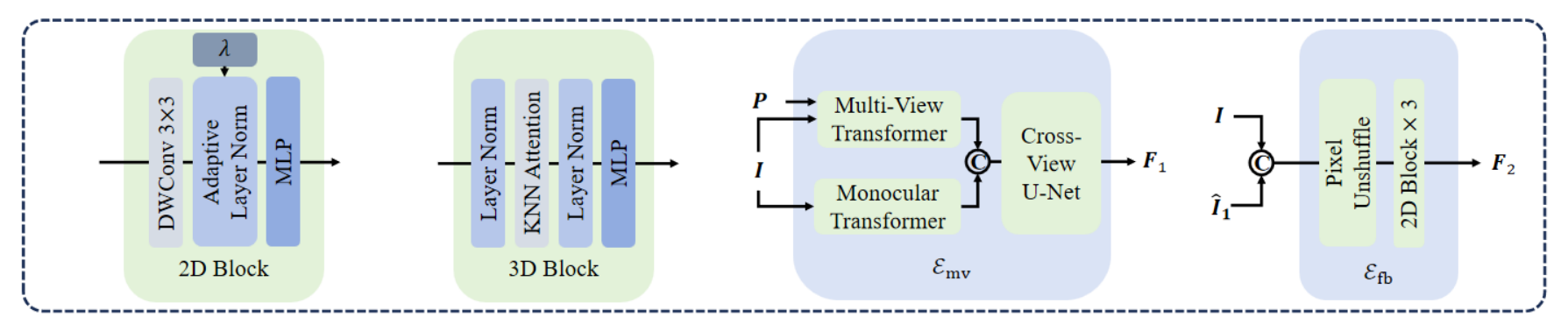}
    \caption{Detailed architectures of the building blocks and feature extractors used in \methodName. The figure illustrates the designs of the 2D and 3D blocks, together with the multi-view feature extractor $\mathcal{E}_{\mathrm{mv}}$ and the feedback-guided feature extractor $\mathcal{E}_{\mathrm{fb}}$.}
    \label{fig:architecture}
\end{figure}

Figure~\ref{fig:architecture} provides the architectural details of the main building blocks and feature extractors used in \methodName. Specifically, the 2D blocks operate on per-view feature maps for intra-view feature transformation, while the 3D blocks aggregate information across views based on the corresponding 3D representations. We further detail the multi-view feature extractor $\mathcal{E}_{\mathrm{mv}}$, which extracts geometry-aware features from the input observations, and the feedback-guided feature extractor $\mathcal{E}_{\mathrm{fb}}$, which jointly processes the original observations and reconstruction feedback to capture complementary information that is not sufficiently represented by the current reconstruction.

\subsection{Details of Multi-View Feature Extraction}

\noindent\textbf{Multi-View Branch.}
Following DepthSplat~\citep{depthsplat}, the multi-view branch first extracts per-view features using a lightweight ResNet-style backbone. The resulting features are subsequently processed by a multi-view Transformer composed of alternating self- and cross-view attention layers, allowing each reference view to aggregate correspondence information from the other input views. For more than two input views, cross-view interaction is restricted to nearby source views following the original DepthSplat design to maintain computational efficiency. We resize the resulting multi-view features to a common spatial resolution of $H/4\times W/4$, yielding
\begin{equation}
\boldsymbol{F}_{\mathrm{mv}}
\in
\mathbb{R}^{N\times C \times \frac{H}{4}\times\frac{W}{4}},
\end{equation}
where $N$ denotes the number of input views and $C$ is the feature dimension.

\noindent\textbf{Cost-Volume Construction.}
To explicitly encode geometric correspondence across views, we construct per-view cost volumes using plane-sweep stereo~\citep{chen2024mvsplat}. For each reference view, a set of depth candidates is sampled within the corresponding near--far range. Features from the source views are warped to the reference camera at each candidate depth using the known camera parameters, and their correlations with the reference features are computed to form a cost volume. When multiple source views are available, the correlations from the selected neighboring views are aggregated to obtain the matching evidence for each reference view.
Different from the original implementation, all feature representations in our framework are ultimately aligned to the $1/4$ spatial resolution before subsequent feature fusion and compression. For RealEstate10K and ACID, the cost volume is directly constructed from the $1/4$-resolution multi-view features. For DL3DV, whose larger spatial resolution and increased number of input views make cost-volume construction substantially more expensive, we first bilinearly downsample the multi-view features to a lower spatial resolution for cost-volume computation and then bilinearly upsample the resulting cost-volume features back to $H/4\times W/4$. This dataset-specific treatment reduces the computational and memory overhead of multi-view matching while keeping the feature resolution presented to the subsequent modules consistent across datasets.

\noindent\textbf{Monocular Branch.}
In parallel with the multi-view branch, the monocular depth branch is initialized from DepthAnything V2~\citep{depthanythingv2} and used to extract per-view geometric priors. Unlike the multi-view features, these monocular features are obtained independently for each view and provide complementary geometric cues in regions where multi-view matching is ambiguous. The monocular features are resized to the same $H/4\times W/4$ spatial resolution before feature fusion.

\noindent\textbf{Cross-View U-Net Fusion.}
Finally, the monocular geometric features, multi-view matching features, and cost-volume information are fused by the cross-view U-Net. In contrast to the original reconstruction pipeline, which progressively upsamples the fused features for depth and Gaussian prediction, we retain the low-resolution U-Net representation at $1/4$ of the input resolution as the output of the multi-view feature extractor. The resulting feature
\begin{equation}
    \boldsymbol{F}_1
\in
\mathbb{R}^{N\times C\times\frac{H}{4}\times\frac{W}{4}}
\end{equation}
serves directly as the input to our first-stage feature codec.

\subsection{Codec and Baseline Settings}
\label{sec:appendix_a}
\noindent\textbf{Codec Configuration.}
For both coding stages, the analysis transform $g_a$ and synthesis transform $g_s$ use a feature dimension of 128 channels throughout the main codec blocks. Given an input feature at $1/4$ of the input spatial resolution, the analysis transform $g_a$ produces a 32-channel latent representation:
\begin{equation}
\boldsymbol{y}_k\in\mathbb{R}^{N\times32\times \frac{H}{8}\times \frac{W}{8}}.
\end{equation}
The entropy model operates on the same 32-channel latent dimensionality and consists of a hyperprior~\citep{balle2018} and a quadtree context model~\citep{dcvc-dc}. The corresponding hyper-latent $\boldsymbol{z}_k$ is further downsampled to $1/32$ of the input spatial resolution.

\noindent\textbf{Backbone and Baseline Settings.}
For our method, we adopt DepthSplat-Small as the multi-view feature extraction backbone. For the competing feed-forward 3DGS pipelines, we use DepthSplat-Small and ReSplat-Small, with the model variants selected to keep their parameter counts as close as possible to that of our method for a more comparable evaluation. 
CodecSplat is configured following its official implementation and is built upon DepthSplat-Base.

\section{Additional Experimental Results}
\label{appendix:exp}

\subsection{Compression Using CodecSplat's Multi-View Feature Extractor}
To enable a more controlled comparison with CodecSplat~\citep{codecsplat}, we conduct an additional experiment on DL3DV by replacing the multi-view feature extractor of \methodName with the feature extraction backbone in CodecSplat. Specifically, we take the intermediate features produced by CodecSplat and first apply PixelUnshuffle to reduce their spatial resolution from full resolution to $1/4$ resolution. The downsampled features are then processed by three 2D blocks to adapt them to our feature representation, yielding the multi-view features $\boldsymbol{F}_1$ used as the input to the subsequent coding stage. All remaining components, including the feedback-guided feature extractor $\mathcal{E}_{\mathrm{fb}}$, analysis and synthesis transforms $g_a$ and $g_s$, state update module $\mathcal{U}$, and Gaussian predictor $\mathcal{P}$, retain the original design of \methodName without architectural modification.

For training, we freeze the pretrained feature extraction backbone of CodecSplat and only optimize  the remaining components of \methodName. These components are initialized from our standard lightweight model and further fine-tuned for 150K iterations with the CodecSplat features. As shown in Figure~\ref{fig:rd-performance}, this scaled variant surpasses CodecSplat in RD performance. These results further demonstrate the scalability of our framework, as a stronger feature extractor can improve compression performance while the lightweight receiver architecture remains unchanged. Moreover, the ability to directly accommodate features from a substantially different and larger feature extractor indicates that our compression framework is not tightly coupled to a specific multi-view backbone or feature representation, demonstrating its compatibility with different feed-forward 3DGS architectures.

\subsection{Comparison with Base Models}
\begin{figure}[h]
    \centering
    \includegraphics[width=\linewidth]{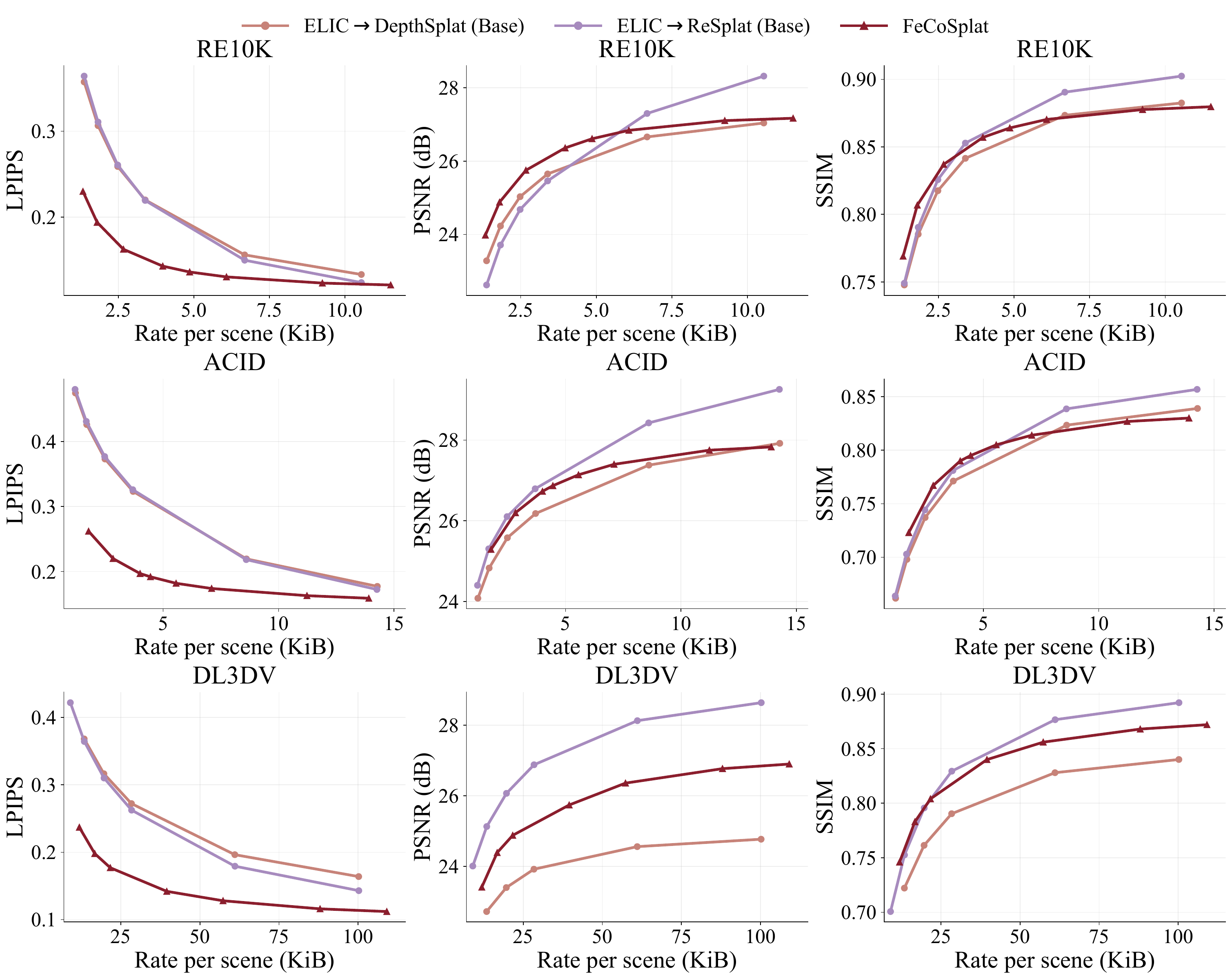}
    \caption{Performance comparison with two base-size methods.}
    \label{fig:base_rd_comparison}
\end{figure}

In the main experiments, model variants with comparable capacities are used whenever available. However, DepthSplat does not provide a corresponding Small model on DL3DV, while ReSplat does not provide Small models on RealEstate10K and ACID. To provide a more comprehensive comparison with the available official models, we additionally compare \methodName with DepthSplat-Base and ReSplat-Base (over 140 M parameters at the receiver). We report the corresponding RD curves and model parameter counts. The results are in Figure~\ref{fig:base_rd_comparison} and Table~\ref{tab:appendix_base_params}. 
Despite its substantially smaller model capacity, \methodName achieves better performance on several metrics at low bitrates and superior LPIPS performance across the evaluated rate range, demonstrating its effectiveness in preserving perceptual quality under constrained bit budgets. Due to time and computation constraints, we are unable to train \methodName at exactly the same model scale as these Base variants. Nevertheless, the scaling experiments in the main paper show that increasing the capacity of \methodName consistently improves reconstruction fidelity, suggesting that the performance gap can be further narrowed with larger models and highlighting the promising scalability of our framework.

\begin{table}[h]
    \centering
    \caption{Comparison of model parameters at the sender and receiver.}
    \vspace{2mm}
    \label{tab:appendix_base_params}
    \begin{tabular}{lcc}
        \toprule
        Methods
        & Sender Params (M) $\downarrow$
        & Receiver Params (M) $\downarrow$ \\
        \midrule

        ELIC$\rightarrow$DepthSplat (Base)
        & 26.45 & 144.81 \\

        ELIC$\rightarrow$ReSplat (Base)
        & 26.45 & 246.20 \\

        \methodName
        & 37.28 & 3.45 \\

        \bottomrule
    \end{tabular}
\end{table}

\newpage

\section{Discussion on Iterative Refinement in Reconstruction and Compression}
\label{appendix:refinement}

Iterative refinement plays a different role in compression from that in reconstruction-oriented methods such as ReSplat~\citep{resplat} and HiSplat~\citep{tang2025hisplat}. In ReSplat, residual updates are directly incorporated into the Gaussian representation without transmission or compression, so additional iterations mainly increase computation and can continuously improve reconstruction quality.

In our framework, each refinement stage is also a coding stage. The complementary information extracted from reconstruction feedback must be quantized, entropy coded, transmitted, and decoded before being integrated into the reconstruction. Therefore, each iteration introduces both lossy compression and additional rate cost, making the gains from repeated refinement smaller from a rate--distortion perspective.

Moreover, our refinement is performed in the compact feature-state domain rather than directly on explicit Gaussian primitives. This preserves the advantages of feature compression while allowing reconstruction feedback to guide subsequent coding. Hence, a small number of refinement stages, such as our two-stage design, provides a better balance between reconstruction fidelity and coding efficiency.

\section{Additional Visualizations}
\label{sec:visualization}

\begin{figure}[h]
    \centering
    \includegraphics[width=\linewidth]{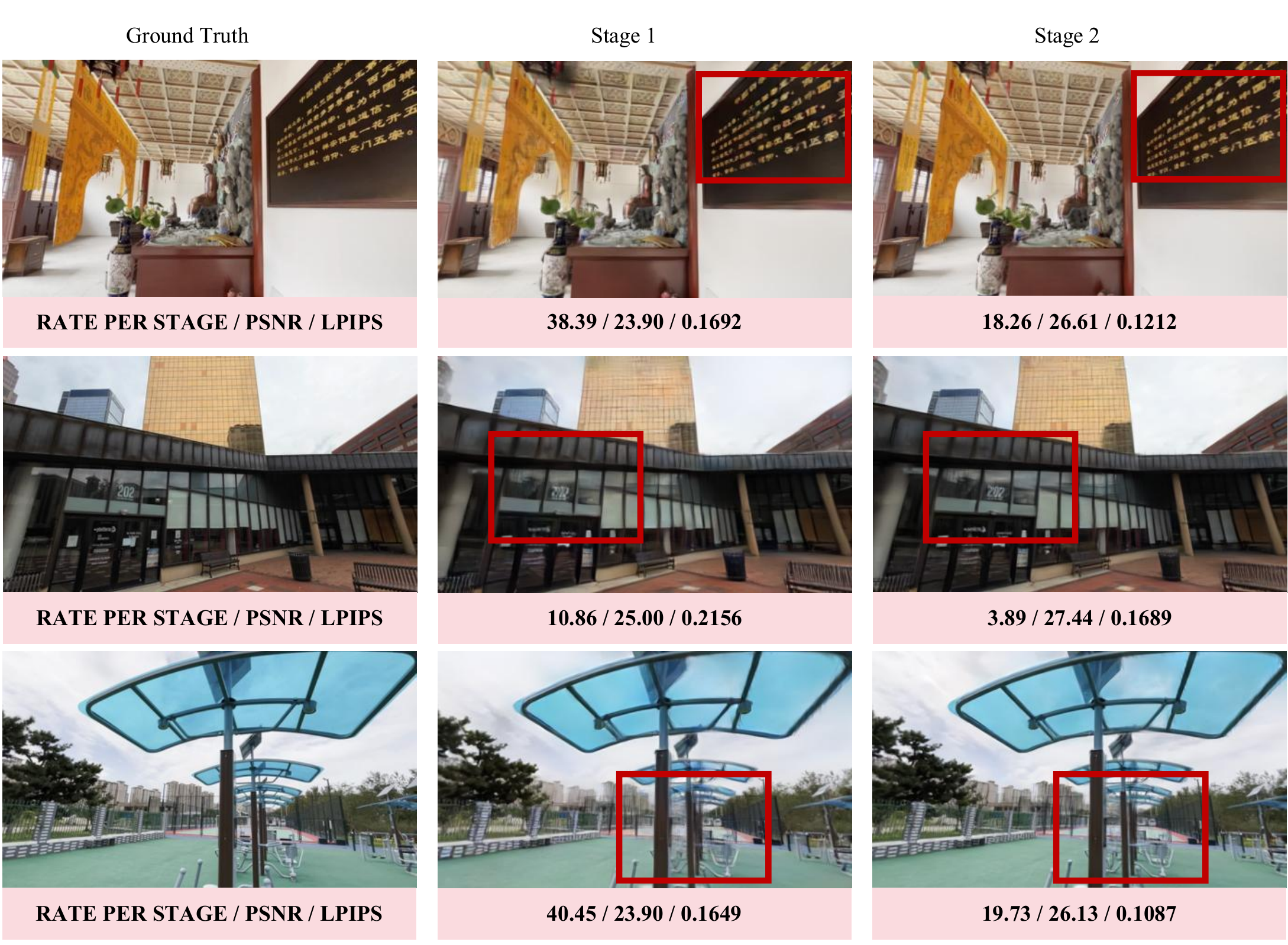}
    \caption{Visualization of the two-stage reconstruction process of \methodName.}
    \label{fig:vis-two-stage}
\end{figure}

\begin{figure}[h]
    \centering
    \includegraphics[width=\linewidth]{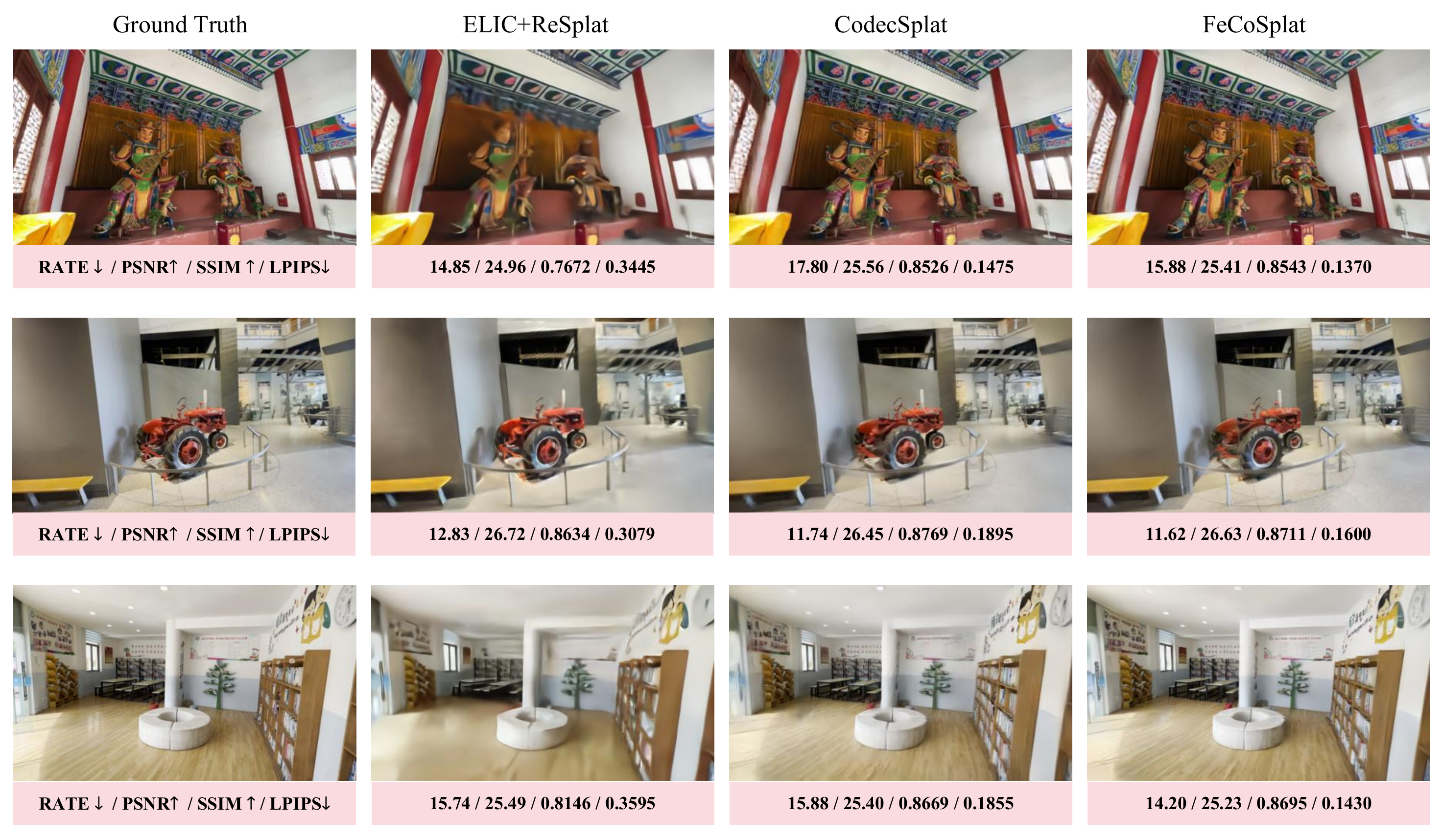}
    \caption{Qualitative comparison on DL3DV.}
    \label{fig:visualization_dl3dv}
\end{figure}

We provide additional qualitative results to further analyze the reconstruction behavior of \methodName. First, we visualize the intermediate and final reconstructions produced by the two coding stages, together with the bitrate contributed by each stage in Figure~\ref{fig:vis-two-stage}, to illustrate how the second-stage feedback-guided features complement the intermediate reconstruction. Although the first-stage Gaussian representation is compact, its rendering quality remains relatively limited, resulting in an unfavorable RD trade-off when used alone. The second stage therefore exploits reconstruction feedback to identify and encode complementary information that is insufficiently captured in the first stage, substantially improving reconstruction fidelity with a moderate additional rate.
Further, we provide qualitative comparisons with competing methods on DL3DV, demonstrating the reconstruction quality of \methodName at low bitrates in Figure~\ref{fig:visualization_dl3dv}.

\newpage

\end{document}